\documentclass{article}

\usepackage{microtype}
\usepackage{graphicx}
\usepackage{subcaption}
\usepackage{booktabs} % for professional tables
\usepackage{adjustbox}

\usepackage{hyperref}
\usepackage{placeins}

\usepackage[preprint]{icml2026}

\usepackage{amsmath}
\usepackage{amssymb}
\usepackage{mathtools}
\usepackage{amsthm}

\usepackage[capitalize,noabbrev]{cleveref}

\theoremstyle{plain}

\theoremstyle{definition}

\theoremstyle{remark}

\usepackage[textsize=tiny]{todonotes}

\icmltitlerunning{Supercharging SBI for BOED}

\begin{document}

\twocolumn[
  \icmltitle{Supercharging Simulation-Based Inference\\ for Bayesian Optimal Experimental Design}



  \icmlsetsymbol{equal}{*}

  \begin{icmlauthorlist}
    \icmlauthor{Samuel Klein}{slac}
    \icmlauthor{Willie Neiswanger}{uca}
    \icmlauthor{Daniel Ratner}{slac}
    \icmlauthor{Michael Kagan}{slac}
    \icmlauthor{Sean Gasiorowski}{slac}
  \end{icmlauthorlist}

  \icmlaffiliation{slac}{SLAC National Accelerator Laboratory}
  \icmlaffiliation{uca}{University of Southern California}

  \icmlcorrespondingauthor{Samuel Klein}{samklein@slac.stanford.edu}

  \icmlkeywords{Machine Learning, ICML}

  \vskip 0.3in
]



\printAffiliationsAndNotice{}  % no special notice (required even if empty)

\begin{abstract}
  Bayesian optimal experimental design (BOED) seeks to maximize the expected information gain (EIG) of experiments. This requires a likelihood estimate, which in many settings is intractable. Simulation-based inference (SBI) provides powerful tools for this regime. However, existing work explicitly connecting SBI and BOED is restricted to a single contrastive EIG bound. We show that the EIG admits multiple formulations which can directly leverage modern SBI density estimators, encompassing neural posterior, likelihood, and ratio estimation. Building on this perspective, we define a novel EIG estimator using neural likelihood estimation. 
  Further, we identify optimization as a key bottleneck of gradient based EIG maximization and show that a simple multi-start parallel gradient ascent procedure can substantially improve reliability and performance. With these innovations, our SBI-based BOED methods are able to match or outperform by up to $22\%$ existing state-of-the-art approaches across standard BOED benchmarks.
\end{abstract}

\section{Introduction}
Modern science experiments are often complex, high dimensional, and expensive. 
To accomplish experimental goals, naive data collection approaches, such as grid or raster scans of spaces of interest, are often intractable, motivating more advanced methods for optimizing data acquisition.
Here the focus is on cases where the scientific goal is to increase information about underlying system parameters, for which Bayesian Optimal Experimental Design (BOED)~\cite{modern_bed} has emerged as the state-of-the-art. 
In BOED, the objective is typically to design a sequence of experiments to maximize the expected information gain (EIG) about parameters of interest. 

The fundamental object in BOED is the likelihood. 
While not all science cases make use of BOED for data acquisition, it is typical for this likelihood to be used in scientific analyses. 
Of particular interest for this work are contexts where explicit likelihoods are not available; in such cases, the likelihood is accessed indirectly via samples generated by a simulator.
% Not enough space here for the things in brackets but leaving as is for .tex viewing
This is common practice in domains such as particle physics~\cite{sbi_atlas}% (inference of theory parameters)
, astrophysics~\cite{Alsing_2019} % (lensing and galaxy simulation pipelines)
and neuroscience~\cite{deistler2022truncated}, % (Hodgkin–Huxley simulations). 
and significant effort has been invested to develop simulators of the underlying processes for each field. 
The burgeoning field of simulation-based inference (SBI)~\cite{cranmer2020frontier} has demonstrated particular success for scientific analysis without explicit likelihoods.

For both explicit and implicit likelihoods, BOED methods provide strategies for finding designs that maximize the EIG.
There are two main approaches that have emerged as promising candidates for optimizing the EIG in high dimensions~\cite{modern_bed}.
\textit{Policy-based} approaches learn an amortized design policy, while \textit{per-trajectory} approaches solve an EIG optimization after each measurement.
In a policy-based approach a network is trained to adapt to any realization of the parameters sampled from the prior.
At deployment, the policy samples designs without further optimization, enabling fast data acquisition.
In contrast, per-trajectory approaches run an EIG optimization procedure after every measurement, deriving solutions specific to each sequence of measurements performed.
Surprisingly, prior work has found that per-trajectory optimization 
performs poorly compared to policy-based strategies~\cite{idad,sbi_bridging}, 
despite the former customizing its design selection to the current set of observations.

This under-performance is especially problematic in settings where simulations are 
expensive and training a policy is intractable, precisely where SBI methods 
are commonly employed~\cite{cranmer2020frontier}.
Further, while SBI has produced useful posteriors across diverse scientific domains, little work has directly connected these methods to BOED, with existing 
attempts reaching a significantly lower EIG than policy-based approaches~\cite{sbi_bridging}.
In the following we outline our contributions to this effort.

\textbf{Comprehensive SBI-BOED framework.} We provide an explicit connection between the three most widely-used SBI methods--neural posterior, likelihood, and ratio estimation--and variational bounds on the EIG.
Specifically, we use each of these SBI methods to directly compute variational approximations to the densities required for relevant formulations of the EIG bounds.
This affords practitioners the flexibility to choose the most appropriate method for their setting.
As a consequence of this connection, we define a \textbf{new estimator for the EIG} and analyze its performance and theoretical properties.

\textbf{Improved EIG optimization.} 
We identify that the underperformance of per-trajectory gradient-based EIG optimization is primarily due to convergence to local optima. 
We propose multiple parallel restart gradient ascent (MPR-GA), a simple algorithm that significantly improves optimization performance, enabling SBI-based methods to match or exceed state-of-the-art policy-based approaches.
The increase in performance is significant, not an incremental improvement, revealing that the upper bound on performance is higher than previously thought.
To enable the MPR-GA technique, we identify and characterize challenges due to amortizing over designs.

We emphasize that, beyond our detailed contributions, there are important practical and conceptual connections between the fields of SBI and BOED. 
Many scientific domains have adopted SBI for posterior inference, leading to extensive development of specialized neural architectures and software tooling~\cite{BoeltsDeistler_sbi_2025}.
Modeling of densities is a dominant focus for SBI, with much effort devoted to flexibility, quality, and inductive biases~\cite{npe_tailored_to_domain}.
BOED, which has historically used relatively simple variational models, has much to gain by leveraging this infrastructure.
Further, by explicitly connecting SBI posteriors to BOED, including describing how to sequentially update SBI posteriors, we expect that many existing scientific workflows may be translatable to data acquisition frameworks.

\section{Background / Related Work}
While there has been work on BOED for implicit models~\cite{mine_bed,idad,efficient_implicit,sbi_bridging}, these approaches do not fully leverage or connect to modern SBI methods. 
The one exception that makes explicit connections to SBI~\cite{sbi_bridging} restricts itself to a single contrastive EIG bound. This leaves a broad range of SBI methodology unexploited for BOED.

\subsection{Simulation based inference}
% Introduction to SBI, what it is, why it's useful
In many scientific domains, simulations provide an accurate model of the process that generated the data, with simulators encompassing all that is known about a system of interest.
Simulators provide a way of sampling observations \(y\), given parameters of interest \(\theta\) and experimental designs \(\xi\), through a likelihood function \(p(y|\theta, \xi)\).
These likelihoods are often intractable, and SBI methods were developed to perform inference in these settings~\cite{cranmer2020frontier}.
SBI methods provide ways of estimating posterior distributions \(p(\theta|D)\) where $D$ denotes previously observed data, using samples from a simulator.
Many SBI methods focus on improving sample efficiency through adaptive simulation schemes~\cite{papamakarios_npe,lueckmann_npe,greenberg_npe,snpea_2021,griesemer2024active,nle_papamakarios}.
These schemes are often essential due to the high computational cost of running simulations in scientific contexts and the high dimensionality of the parameter space.

This work aims to pave the way for existing SBI infrastructure to be used directly for data acquisition using BOED.
Many scientific workflows have already invested in SBI infrastructure for posterior inference.
This involves specialized simulators, training pipelines, and validation frameworks~\cite{hermans2022a,lueckmann2021benchmarking}.
These SBI systems often provide amortized posteriors that enable on-the-fly inference without retraining.
In this work we will assume that all simulators are differentiable. This makes gradient-based optimization of the EIG more straightforward, as gradients can be backpropagated through the simulator. Differentiable simulators are becoming more common in scientific contexts~\cite{diffable_climate,neutrino-diffable,MadJax,Neos} due to an increased understanding of automatic differentiation~\cite{JMLR:v18:17-468}. We leave the case of non-differentiable simulators for future work and provide further discussion in Appendix~\ref{app:differentiable_simulators}.

\subsection{Bayesian optimal experiment design}
BOED~\cite{modern_bed} is a framework for selecting experimental designs \(\xi\) that maximize the EIG about \(\theta\).
The EIG for a candidate design \(\xi\) quantifies the expected reduction in entropy of parameters \(\theta\) from observing data \(y\), and admits two equivalent formulations:
\begin{align}
  \mathrm{EIG}(\xi) &= \mathbb{E}_{p(\theta|D)p(y| \theta, \xi)} \left [ \log \frac{p(\theta|y, \xi)}{p(\theta|D)} \right ] \label{eq:npe_acq}\\
  &= \mathbb{E}_{p(\theta|D)p(y| \theta, \xi)} \left [ \log \frac{p( y | \theta, \xi)}{p(y|\xi)} \right ] \label{eq:nle_acq}.
\end{align}
Even when the likelihood $p(y|\theta, \xi)$ is tractable, the posterior $p(\theta|y, \xi)$ and marginal likelihood $p(y|\xi)$ are often intractable.
A straightforward Monte Carlo approach requires nested sampling, making the estimator prohibitively inefficient~\cite{rainforth2018nesting}, especially in high dimensional settings.
An alternative approach is to use variational bounds.
Several types of bounds have been developed for the EIG, including lower bounds that can be jointly optimized for variational parameters and designs~\cite{foster_unified}.
% Split?
The focus of this work is on per-trajectory strategies, where the posterior is updated and the EIG is optimized with respect to the current posterior after each measurement.
We consider in particular expensive measurement processes where significant computational resources can be applied to select the most informative designs.
This includes settings where there are minutes to hours between measurements, such as in astronomy or materials/beamline experiments.

\section{Methods}
In this section, we describe how to integrate SBI methods into BOED using gradient-based optimization of the EIG. In particular, we propose one new EIG estimator and outline how to formulate and optimize three existing EIG bounds using our approach. For the following, we assume that the designs $\xi$ at which the EIG is computed are drawn from a design distribution $p(\xi)$. 
We construct the variational approximations needed to optimize the EIG using SBI methods, which are designed to estimate these densities in the context of implicit likelihoods. 
Importantly, different SBI approaches naturally correspond to different EIG lower bound formulations. 

\subsection{Neural likelihood estimation}
As a first example of the utility of our methodological framework, we consider neural likelihood estimation (NLE).
In the NLE formulation, the likelihood function \(p(y|\theta, \xi)\) is approximated using a conditional density estimator \(q_\phi(y|\theta, \xi)\) with parameters \(\phi\)~\cite{nle_papamakarios}.
This density estimator is trained using maximum likelihood on samples drawn from the joint distribution \(p(\theta, y, \xi) = p(y|\theta, \xi)p(\theta|D)p(\xi)\).
Samples from the posterior distribution can then be obtained using Bayes' rule.
There are several approaches to constructing this density estimator, including the use of normalizing flows~\cite{tabak2013family,rezende2015variational} which are popular in the SBI literature.

\textbf{Estimator 1: new, direct EIG estimator with NLE.} The EIG formulation that most naturally aligns with NLE uses the likelihood \(p(y|\theta, \xi)\) and marginal likelihood \(p(y|\xi)\) as in Eq.~\eqref{eq:nle_acq}.
We notice that both densities share the same observation space, so we use the same neural architecture to model each, giving
\begin{align}
  \mathrm{EIG}(\xi) \approx \mathbb{E}_{p(\theta|D)p(y| \theta, \xi)} \left [ \log \frac{q_\phi( y | \theta, \xi)}{q_\phi(y|\xi)} \right ] \label{eq:nle_approx},
\end{align}
where separate normalizing flows parameterize each density.
While conceptually straightforward, we are not aware of prior applications of this acquisition function in BOED.
This estimator requires training two models. We train the estimator \(q_{\phi}(y|\xi)\) using the same training set of parameter-observation-design-data pairs as we use for \(q_{\phi}(y|\theta, \xi)\).
An analysis of our approach is provided in Appendix~\ref{app:direct_eig_estimate}.

\textbf{Estimator 2: Contrastive estimator with NLE.} A less direct approach that uses NLE is to construct a contrastive estimator for the EIG~\cite{sbi_bridging}.
This estimator only requires a single estimator, for the marginal likelihood, and was shown to perform poorly on the EIG compared to policy-based approaches. Our results suggest that this poor performance was due to optimization and can be addressed with our innovations (Appendix~\ref{app:acq_opt}). However, the disadvantage of the contrastive bound is that it requires many samples from the prior to obtain a low variance estimate of the EIG.
This can be computationally expensive, especially in high dimensional parameter spaces with complex normalizing flow-based likelihood estimators.
This is in contrast to the estimator in Eq.~\eqref{eq:nle_approx}, where there is no additional expectation over samples from the prior. We therefore proceed with Eq.~\eqref{eq:nle_approx} as our nominal NLE formulation of the EIG.

\subsection{Neural posterior estimation}
% Introduction to NPE, what it is, why it's useful
In contrast to NLE, which estimates the likelihood, neural posterior estimation (NPE) directly approximates the posterior \(p(\theta|\xi, y)\) using a neural network~\cite{papamakarios_npe}.
This is done by training a conditional density estimator \(q_\phi(\theta|\xi, y)\), with parameters \(\phi\), to minimize the Kullback-Leibler divergence between the true posterior and the estimated posterior.
As with NLE, normalizing flows are often used to estimate the posterior distribution.

% Connection between NPE and EIG estimation
\textbf{Estimator 3: Barber-Agakov bound with NPE.} Using the definition of the EIG in Eq.~\eqref{eq:npe_acq}, we replace the true posterior \(p(\theta|\xi, y)\) with the variational posterior \(q_\phi(\theta|\xi, y)\) to obtain a lower bound on the EIG as
\begin{equation}
  \mathrm{EIG}(\xi) \geq \mathbb{E}_{p(\theta|D)p(y| \theta, \xi)} \left [ \log q_\phi(\theta|y, \xi) \right ].
  \label{eq:ba_bound}
\end{equation}
This connection between NPE and the EIG is called the Barber-Agakov bound~\cite{barber2004algorithm} and has been explored in the context of BOED in prior work~\cite{foster_variational}.
To date, flexible variational estimators such as normalizing flows have not been well explored in the BOED context, and we address this gap.

\subsection{Neural ratio estimation}

\textbf{Estimator 4: InfoNCE bound with NRE.} Another way to perform SBI is to estimate the likelihood-to-evidence ratio directly using a classifier~\cite{thomas2022likelihood}.
This version of SBI has been shown to be consistent with contrastive learning approaches and is commonly referred to as neural ratio estimation (NRE)~\cite{contrastive_durkan}.
Embedding this approach into BOED naturally motivates the use of contrastive bounds on the EIG.
Contrastive bounds have been explored for both per-trajectory EIG optimization~\cite{efficient_implicit,mine,mine_bed} and policy-based approaches~\cite{idad}.
In particular, the InfoNCE bound~\cite{oord2019representationlearningcontrastivepredictive} has demonstrated strong performance in the policy-based setting~\cite{idad}; we apply it to per-trajectory optimization for the first time.

In general, we present the natural correspondence between the form of the SBI density estimator and the form of the EIG bound as an advantage, and our nominal approach uses each EIG estimator with its corresponding SBI technique. 
However, it is possible to mix components. 
We found that using NPE posteriors with the InfoNCE bound--the EIG formulation associated with NRE--can be advantageous for both model quality and speed.
We refer to this hybrid as NPE-NRE, while methods using the same technique for both components are denoted by a single name, such as NRE.
The key advantage is computational: NRE requires expensive MCMC for posterior sampling, whereas NPE provides samples directly from a normalizing flow.
Similarly, NPE samples can be used with our NLE estimator from Eq.~\eqref{eq:nle_approx}, avoiding NLE's MCMC requirement.
However, since the NLE EIG estimate evaluates two density models, it remains more expensive than NPE-NRE.

\subsection{Multiple parallel restart gradient ascent} 

\begin{algorithm}[t]
\caption{MPR-GA for SBI-BOED. The EIG estimator \(\widehat{\mathrm{EIG}}_\phi(\xi)\) can be defined using any of the SBI based approaches described in this work.}
\label{alg:mpr_ga}
\begin{algorithmic}[1]
\REQUIRE Current posterior $p(\theta|D_t)$, design distribution $p(\xi)$, simulator $p(y|\theta,\xi)$
\REQUIRE Hyperparameters: $M$ restarts, $K$ gradient steps, learning rate $\alpha$, batch size $S$

\STATE Sample buffer $\Theta \sim p(\theta|D_t)$
\STATE Initialize designs: $\{\xi_j^{(0)}\}_{j=1}^M$ where $\xi_j^{(0)} \sim p(\xi)$
\FOR{$k = 1, \ldots, K$}
    \FOR{$j = 1, \ldots, M$ \textbf{(parallel)}}
        \STATE Sample $\{\theta_i\}_{i=1}^S \sim \Theta$
        \STATE Sample $\{y_i\}_{i=1}^S$ where $y_i \sim p(y|\theta_i, \xi_j^{(k-1)})$
        \STATE Update $\phi$ with SGD \COMMENT{Adapt to current designs}
        \STATE $g_j \gets \nabla_\xi \widehat{\mathrm{EIG}}_\phi(\xi_j^{(k-1)})$ on $\{(\theta_i, y_i, \xi_j^{(k-1)})\}_{i=1}^S$
        \STATE $\xi_j^{(k)} \gets \text{proj}_{\Xi}\big(\xi_j^{(k-1)} + \alpha \cdot g_j\big)$
    \ENDFOR
\ENDFOR

\STATE \textbf{return} $\xi^* \gets \arg\max_j \widehat{\mathrm{EIG}}_\phi(\xi_j^{(K)})$
\end{algorithmic}
\end{algorithm}

Previous works have consistently reported that policy-based approaches outperform per-trajectory gradient-based optimization of the EIG, independent of the variational bound that is used~\cite{idad,sbi_bridging,dahlke2025flowbased}.
One hypothesis for this observation is that the variational models used in gradient-based optimization are not sufficiently flexible to capture the true posterior.
However, as seen in previous work~\cite{idad}, when using SBI approaches that accurately estimate posteriors in high dimensions, we observe that policy-based approaches still outperform simple gradient based optimization.
We demonstrate that improved EIG optimization is necessary for per-trajectory SBI-BOED to be competitive with policy-based approaches and provide a procedure for this improvement, namely multiple parallel restart gradient ascent (MPR-GA).

The MPR-GA procedure consists of a robust EIG optimization strategy coupled with an online model update rule and is summarized in Algorithm~\ref{alg:mpr_ga}. Robust optimization is needed as the EIG is generally a non-convex function of the design \(\xi\). To address this, we propose initializing multiple candidate designs \(\{\xi_j\sim p(\xi)\}_{j=1}^M\) and performing gradient ascent on each design in parallel.
After a fixed number of gradient steps, the design with the highest EIG is selected as the optimal design for the next experiment.
This approach allows for exploration of the design space and helps to avoid local optima.

To guarantee EIG estimation quality near relevant designs over the course of the MPR-GA optimization, the model parameters $\phi$ are updated online via stochastic gradient descent (SGD) on batches of simulated data during each iteration of stochastic gradient ascent on the design parameters $\xi$.
The optimization of $\xi$ and $\phi$ is performed in separate steps, contrasting with joint optimization procedures that update both simultaneously~\cite{foster_unified}.
By separating these updates, we leverage the well-established stability and convergence properties of SGD for machine learning models.
This approach also allows for fair comparisons across different acquisition optimization strategies.
This separated strategy has been shown to be useful when building local generative surrogates for detector design problems~\cite{michael_local}.

The MPR-GA algorithm exploits several key properties that make it computationally efficient in practice.
First, the EIG naturally decomposes over designs, meaning gradients for each candidate design $\xi_j$ can be computed independently.
Second, the variational models used for EIG estimation are amortized over the design space, enabling efficient computation of all gradients at arbitrary design points.
Third, modern hardware allows parallel evaluation of multiple designs simultaneously, often without increasing wall-clock time or requiring the allocation of additional compute resources beyond what would be needed for a single design optimization.
Finally, many scientific simulators are embarrassingly parallel, allowing the generation of synthetic data for all $M$ candidate designs concurrently.
Together, these properties enable MPR-GA to explore $M$ different regions of the design space with high parallelization efficiency: information gain increases rapidly with $M$ before plateauing, while wall-clock time remains nearly constant for much of this regime due to parallel execution. While the MPR-GA approach may increase the number of simulations needed to perform the EIG optimization, it leads to significant improvements in the quality of the selected designs.
In settings where real experiments are expensive to run, the additional computational cost is expected to be justified by the improved performance.

A key challenge with MPR-GA is that the design distribution $p(\xi)$ evolves during optimization.
As candidate designs move via gradient ascent, they may cluster in certain regions or fail to explore high-EIG areas where the model has limited training data.
Since simulations are generated at the current design locations, the model's accuracy in any region depends on the density of $p(\xi)$ in that region.
This can cause issues if few designs venture into high-EIG regions, as the model may not correctly identify these areas as valuable.

To address this, we introduce a diversity penalty that encourages designs to spread across the space as
\begin{equation}
\mathcal{L}_{\text{div}}(\boldsymbol{\xi}) = w(t) \sum_{i < j} \max(0, d_{\min} - \|\boldsymbol{\xi}_i - \boldsymbol{\xi}_j\|_2)^2,
\label{eq:desing_penalty}
\end{equation}
where $d_{\min}$ sets a minimum separation threshold, $\|\cdot\|_2$ denotes Euclidean distance, and $w(t)$ is annealed to zero after a fixed number of steps $t$.
By preventing collapse to a single design, this penalty maintains more uniform coverage and helps ensure the model remains performant across the explored regions.
The initial distribution $p(\xi)$ should also provide reasonable coverage of the design space, and we will explore the impact of this selection on the MPR-GA performance.
Despite these challenges, we find the method to be remarkably robust in practice. The simple diversity penalty, combined with reasonable initialization, maintains sufficient model quality throughout optimization across all benchmarks we consider without requiring careful tuning.

\subsection{Sequential SBI}
To compute the EIG at a design $\xi$, samples from the distribution \(p(\theta|D)\) are required.
Before making any measurements this is just the prior, but after every measurement it is replaced by the current posterior.
This density does not need to be known explicitly, only samples from it are needed.
All posterior estimates will be derived from SBI methods that have been adapted to the sequential data acquisition setting.
While SBI methods have been extensively explored in different scientific contexts, their application to sequential data acquisition--where the posterior is updated after each measurement--has several aspects that have not been considered.

We identify two approaches for updating the posterior in sequential settings.
The first approach is to take the current posterior $p(\theta|D)$ as the prior for the next round of inference and define a conditional density estimator that is conditioned only on the latest observation and design pair.
The second approach is to train a new conditional density estimator that is conditioned on all previous observations and designs as well as samples from the prior $p(\theta)$ or a different proposal distribution $\pi(\theta)$.
The choice between these approaches depends on the SBI method used.
Previous work has solved the design optimization and posterior update jointly~\cite{mine_bed}, but we note that this falls strictly under the first approach and does not leave flexibility to choose between the two.
The joint approach can lead to biased designs due to training only on samples from the current posterior.
To avoid this issue we separate the design optimization and posterior update steps.

For NLE and NRE methods, we train variational models conditioned on all measured design points, as in the second approach.
This allows us to use straightforward Markov Chain Monte Carlo (MCMC) with the Metropolis-Hastings algorithm~\cite{metropilis_hastingupdated} to generate samples from the posterior.
While both methods can in principle be trained on samples from any proposal distribution $\pi(\theta)$, including the current posterior, we found this can lead to bias. 
Instead, we use an even mixture of samples from the current posterior and the prior.
% Split?
% Sequential NPE and how it is done
For NPE, we adopt the first approach, taking the current posterior estimate as the prior for the next round of inference.
This is similar to existing adaptive simulation schemes~\cite{lueckmann_npe,greenberg_npe} that refine the proposal distribution over parameters.
In each new round, we train a conditional density estimator \(q_\phi(\theta|\xi, y)\) to approximate the posterior given only the latest observation and design pair.
This approach requires only a single density estimator to be trained at each round, reducing computational cost.
Adaptive simulation schemes can be used to further improve simulation efficiency in this setting.

\section{Results}
For all results in this section the simulators are treated as implicit models where the likelihood function is intractable and can only be accessed via samples. We use as the primary metric the sequential Prior Contrastive Estimation (sPCE) bound~\cite{dad}, which provides a lower bound on the total information gained from a sequence of measurements.
This bound is standard in the BOED literature.
In the following we first demonstrate the impact of the MPR-GA procedure and then compare to state-of-the-art baselines.
We will primarily compare our approaches to deep adaptive design (DAD)~\cite{dad} and reinforcement learning BOED (RL-BOED)~\cite{rl_boed}.
Both of these methods are state-of-the-art policy-based approaches that have consistently outperformed per-trajectory techniques.
Unless otherwise stated, all MPR-GA procedures are run for $K=12000$ steps with $S=3$ samples.
The InfoNCE bound is always estimated with 1024 contrastive samples.
All flows are constructed from rational quadratic spline layers~\cite{durkan2019neural}. For all model details see Appendix~\ref{app:hyperparameters}.
A full breakdown of computational costs is provided in Appendix~\ref{app:timing_information}.
All code is available at \href{https://github.com/sambklein/sbi-boed}{github.com/sambklein/sbi-boed}.

\subsection{Improving acquisition optimization}
\label{sec:ablations}
We first evaluate MPR-GA on single-round data acquisition, where one design is selected based on the current posterior.
As this is computationally inexpensive, we can evaluate this scenario many times to get an accurate estimate of the sPCE bound.
We consider the problem of locating two sources in 2D space (four total parameters to be learned) using noisy sensor measurements~\cite{dad}. 
Each data acquisition step consists of placing a sensor at design location $\xi$ and recording the summed intensity from both sources, including noise. 
The intensity from each source decays with the square of the distance from the source to the sensor. 
This setting is selected because it produces posteriors with structure, has a continuous design space requiring gradient based optimization, and results in multiple local minima that challenge optimization methods.
To explore settings with interesting posterior structure, observations are made at two different locations simultaneously; one point is selected to be at the origin and an additional point is selected by randomly selecting one of the two sources and then sampling uniformly from a ball of radius 0.5 around that source.

%  Plots are made with plot_posterior_and_opt.py in sbi_bax repo
\begin{figure*}[htbp]
    \centering
    \begin{subfigure}[b]{0.28\textwidth}
        \centering
        \includegraphics[width=\textwidth]{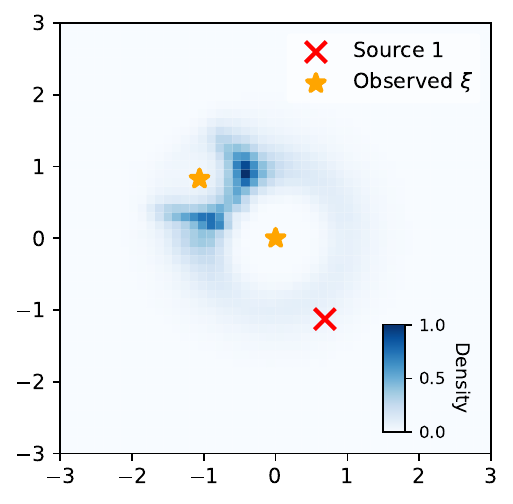}
    \end{subfigure}
    \hfill
    \begin{subfigure}[b]{0.28\textwidth}
        \centering
        \includegraphics[width=\textwidth]{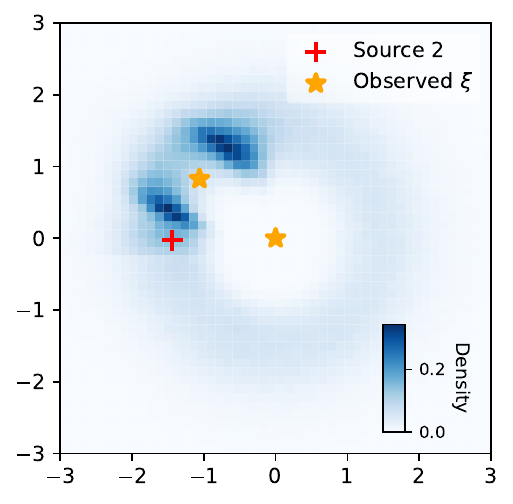}
    \end{subfigure} 
    \hfill
    \begin{subfigure}[b]{0.28\textwidth}
        \centering
        \includegraphics[width=\textwidth]{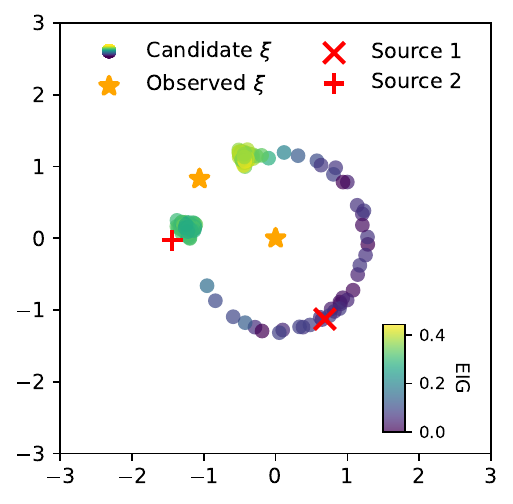}
    \end{subfigure}
    \caption{Posterior samples for the first (left) and second (center) sources after making observations at two design points. The observed designs $\xi$, the source locations and the distribution of candidate designs after running the MPR-GA acquisition optimization with 256 restarts for 12000 steps for the NRE method (right).}
    \label{fig:source_finding_posterior}
\end{figure*}
After making two observations, a posterior is defined using NPE, and each EIG formulation is used to select a design.
The posterior is defined using NPE because it allows for efficient sampling from the posterior using normalizing flows, keeping the full procedure computationally inexpensive.
As all EIG formulations use the same posterior samples, this choice does not favor any particular method.
An example posterior and corresponding optimization is shown in Figure~\ref{fig:source_finding_posterior}.
This figure highlights the need for multiple restarts in the gradient ascent procedure, as many candidate designs get stuck in poor local optima, and the EIG surface is flat in many regions of $\xi$.

\begin{figure*}[htbp]
    \centering
    \begin{subfigure}[b]{0.29\textwidth}
        \centering
        \includegraphics[width=\textwidth]{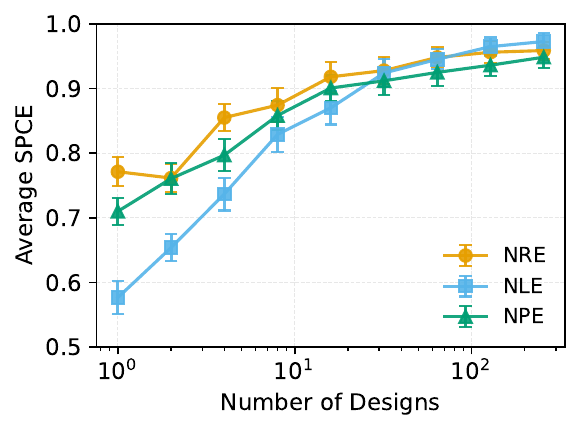}
    \end{subfigure}
    \hfill
    \begin{subfigure}[b]{0.29\textwidth}
        \centering
        \includegraphics[width=\textwidth]{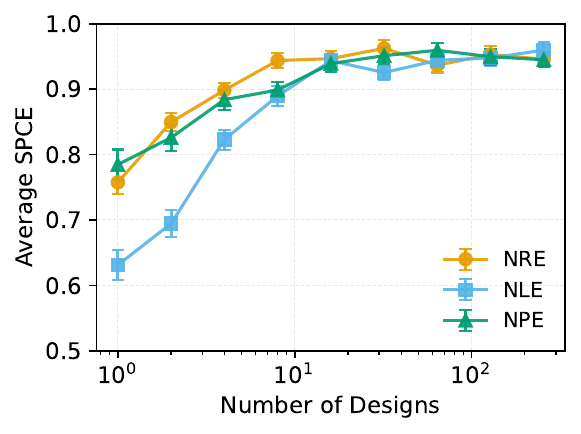}
    \end{subfigure}
    \hfill
    \begin{subfigure}[b]{0.29\textwidth}
        \centering
        \includegraphics[width=\textwidth]{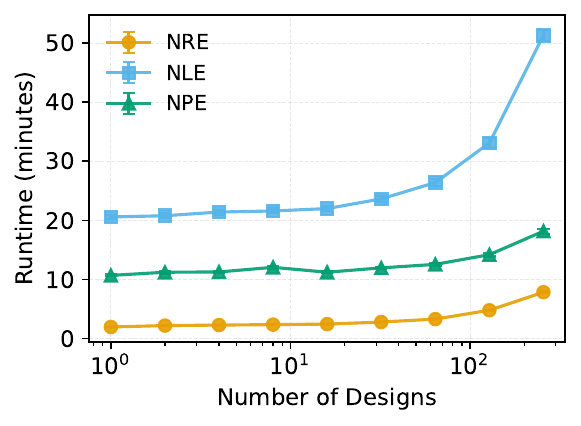}
    \end{subfigure}
    \caption{Ablation on the number of restarts used in MPR-GA for different SBI methods. Using multiple restarts significantly improves performance across all SBI methods considered. Error bars show standard error across 50 random seeds. Different initialization strategies are considered: random designs (left) and sampling from the current posterior (middle). The timing of each method is also shown (right).}
    \label{fig:ablation_restarts}
\end{figure*}

We explore the impact of the number of designs used in the MPR-GA strategy as well as the strategy used to initialize the design points.
Our default strategy for initialization is to sample designs randomly from a given distribution $p(\xi)$.
Previous work on source finding has generated random designs by drawing samples for single sources individually from the prior; this is taken to be the default random design initialization.
However, an alternative strategy is to sample designs using the current posterior.
We clearly see that using multiple restarts significantly improves performance across all EIG formulations considered, as shown in Figure~\ref{fig:ablation_restarts}.
We also see that the initialization strategy has a significant impact: initializing the optimization with samples from the current posterior saturates performance with fewer restarts.
Importantly, the computational time required for each method grows slowly with the number of restarts, allowing for significant performance gains with minimal additional computational cost.

Using the MPR-GA procedure with the diversity penalty of Eq.~\eqref{eq:desing_penalty} improves performance across all SBI methods considered (Appendix~\ref{app:acq_opt}).
All optimization in the following results uses the diversity penalty.
The NLE method using a contrastive bound for the EIG performs similarly to the other methods when using MPR-GA and 1024 contrastive samples, but is approximately $6$ times slower than the formulation that uses Eq.~\eqref{eq:nle_approx} due to the large number of prior samples required, so we choose to not include it in our benchmarks.
See Appendix~\ref{app:acq_opt} for further details on the diversity penalty and contrastive NLE.
All methods were benchmarked on a single NVIDIA RTX 2080 Ti GPU.

\subsection{Source finding in $n$ dimensions}
To compare to state-of-the-art policy-based approaches we turn to the full problem of locating the positions of two sources in a 2D, 3D, and 5D space.
This means that the parameter dimension is 4, 6, and 10 respectively.
We consider data acquisition over ten rounds.
Each SBI method is first benchmarked following the correspondence between posterior estimation and EIG formulations we identify above, as shown in Table~\ref{tab:source_finding_results}.
For our benchmarks, we find that all SBI methods perform similarly, but NPE is significantly faster as a posterior estimator and the NRE EIG formulation is significantly faster to compute for acquisition.
As the most computationally efficient of our approaches, we focus later experiments on this combined NPE-NRE method. However, we emphasize that this NPE-NRE advantage is specific to our benchmarking scenario, and, in practice, performance of each method will depend strongly on particular simulation design and infrastructure. 

To date the best published results on the source finding benchmark were achieved using DAD*~\cite{idad}, a policy-based approach which samples designs that maximize the EIG.
When applying the MPR-GA procedure with 256 restarts, we find that all three SBI-BOED methods significantly outperform DAD in 2D source finding and perform similarly in the other considered dimensionalities, as shown in Table~\ref{tab:source_finding_results}.
We also find that MPR-GA significantly outperforms gradient-based optimization approaches using only a single design initialization~\cite{efficient_implicit,mine,mine_bed} across all dimensions considered. In particular, while $M=1$ NRE and MINEBED* perform similarly, $M=256$ NRE demonstrates a large gain, which comes from this improved optimization. 
We stress that our improvements are a significant leap in performance over previous per-trajectory approaches.
The fact that per-trajectory methods excel in 2D but show diminishing advantages in higher dimensions suggests posterior estimation quality degrades in higher dimensions, hampering per-trajectory optimization, and that more MPR-GA candidate designs may be required.

\begin{table}[h]
\centering
\caption{Average sPCE lower bound across for source finding with different acquisition methods and dimensions. Error shown as standard error. Comparison numbers for DAD* and MINEBED* are taken from the iDAD paper~\cite{idad} where this model performed best. All SBI-BOED methods are averaged over 10 runs except for NPE-NRE which is averaged over 100 runs, all other methods use 4096 runs.}
\adjustbox{max width=\columnwidth}{
\begin{tabular}{lccc}
\toprule
Method & 2D & 3D & 5D \\
\midrule
Random & $4.79 \pm 0.04$ & $3.47 \pm 0.01$ & $1.89 \pm 0.01$ \\
\midrule
\multicolumn{4}{l}{\textit{Policy-based}} \\
DAD* & $7.97 \pm 0.03$ & $6.30 \pm 0.03$ & $3.34 \pm 0.04$ \\
RL-BOED & $6.86 \pm 0.05$ & $4.95 \pm 0.05$ &--\\
\midrule
\multicolumn{4}{l}{\textit{Per-trajectory M = 1 restart}} \\
MINEBED* & $5.52 \pm 0.03$ & $4.22 \pm 0.03$ & $2.46 \pm 0.03$ \\
NRE & $5.67 \pm 0.59$ & $4.39 \pm 0.85$ & $2.58 \pm 0.79$ \\
\midrule
\multicolumn{4}{l}{\textit{Per-trajectory M = 256 restarts}} \\
NRE & $7.41 \pm 0.64$ & $6.44 \pm 0.92$ & $3.07 \pm 0.87$ \\ 
NLE & $\mathbf{9.62 \pm 0.82}$ & $5.37 \pm 0.76$ & $3.01 \pm 0.86$ \\
NPE & $9.12 \pm 0.99$ & $5.08 \pm 0.97$ & $\mathbf{3.62 \pm 0.60}$ \\
NPE-NRE & $9.27 \pm 0.31$ & $\mathbf{6.44 \pm 0.28}$ & $3.13 \pm 0.20$ \\
\bottomrule
\end{tabular}
}
\label{tab:source_finding_results}
\end{table}

\subsection{Pharmacokinetic model}
The next benchmark we consider is a pharmacokinetic model~\cite{idad,ryan2014towards}.
The pharmacokinetic model involves selecting timepoints for drug concentration measurements to best infer pharmacokinetic parameters.
This is a 1D design problem where policy-based methods should excel due to their fast deployment.
The design distribution $p(\xi)$ is taken to be uniform across the design space.

We find that per-trajectory optimization discovers highly consistent designs across the first three measurements. 
The NPE-NRE approach frequently selects measurements at approximately the same three timepoints across different experimental runs.
Using these timepoints as a static design baseline outperforms both the DAD and the RL-BOED approaches across three measures as shown in Table~\ref{tab:pharmacokinetic_results}, even when both of the policies from these methods are retrained to only perform three measurements.
Notably, this static baseline requires zero deployment time, providing both better performance and faster execution than the trained policy.

This result challenges the central advantage of policy-based methods---fast deployment---and suggests that improved policy training strategies may be needed even in settings where policies should excel.
We note that the advantage of the SBI per-trajectory approach diminishes as more measurements are made, with DAD performing similarly to NPE-NRE after five measurements.
For a static baseline over five measurements we find repeating two of the three timepoints used in the three measurement static baseline works very well, again suggesting that strong static baselines may exist for this problem.

\begin{table}[h]
\centering
\caption{Average sPCE lower bound for the pharmacokinetic model after $T$ measurements. Errors show standard error over experimental runs. For the NPE-NRE baseline 100 runs were performed, for all other methods 4096 runs were used including the static baseline.}
\adjustbox{max width=\columnwidth}{
\begin{tabular}{lcc}
\toprule
Method & $T=3$ & $T=5$ \\
\midrule
Random & $2.14 \pm 0.02$ & $2.75 \pm 0.03$ \\
DAD & $2.12 \pm 0.02$ & $\mathbf{3.23 \pm 0.02}$ \\
RL-BOED & $2.37 \pm 0.03$ & $2.98 \pm 0.03$ \\
NPE-NRE & $2.50 \pm 0.14$ & $3.20 \pm 0.13$ \\
NPE-NRE (static) & $\mathbf{2.56 \pm 0.02}$ & $3.13 \pm 0.02$ \\
\bottomrule
\end{tabular}
}
\label{tab:pharmacokinetic_results}
\end{table}

\subsection{Constant elasticity of substitution}
\begin{figure}[htbp]
    \centering
    \includegraphics[width=0.4\textwidth]{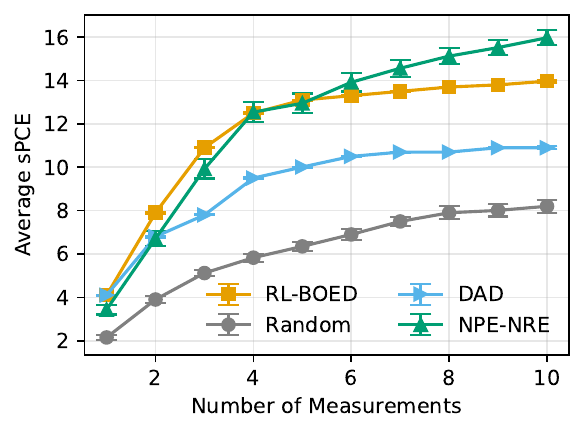}
    \caption{Average sPCE lower bound after 10 measurements for CES experiments. Errors show standard error on the mean over different runs. For NPE-NRE 100 runs were performed, for all other methods 4096 runs were used.}
    \label{fig:ces_plot}
\end{figure}

For our final benchmark we consider the constant elasticity of substitution (CES) production function model~\cite{foster_unified}.
In this setting the design space is six dimensional and the parameters of interest are five dimensional.
In each trajectory ten measurements are performed, and we find that 36000 gradient updates are required for convergence of the MPR-GA procedure.
We see that NPE-NRE substantially outperforms existing methods, achieving an EIG of $15.97 \pm 0.34$ compared to $13.97 \pm 0.06$ for RL-BOED and $10.77 \pm 0.08$ for DAD.
Notably, NPE-NRE reaches the peak performance of RL-BOED after only 6 measurements rather than 10, demonstrating both superior final performance and faster information acquisition. This is shown in Figure~\ref{fig:ces_plot}.
Interestingly, the per-trajectory approach underperforms in early measurements but surpasses policy-based methods later, suggesting that improved posterior estimation quality over time enables more effective design selection.
The design distribution $p(\xi)$ is taken to be uniform across the design space.

\section{Conclusion}
We have demonstrated that modern SBI methods, combined with improved optimization, enable per-trajectory BOED to match or exceed state-of-the-art policy-based approaches and significantly outperform all previous per-trajectory approaches.
Our framework connects all three major SBI approaches (NPE, NLE, NRE) to variational EIG bounds, enabling practitioners to leverage existing SBI infrastructure directly for experimental design.
A key insight is that prior underperformance of per-trajectory methods stems from optimization failures.
Our MPR-GA procedure---combining parallel restarts with online model adaptation---achieves up to 22\% improvement over previous state-of-the-art on source finding in two dimensions and 14\% gains over the best policy-based method on the CES benchmark.
Remarkably, static designs derived from per-trajectory optimization can outperform trained policies while requiring zero deployment time, challenging assumptions about the necessity of policy amortization.

In higher-dimensional settings, our advantage over policy-based methods diminishes, and it is likely that posterior estimation quality is limiting acquisition effectiveness.
This highlights the need for SBI methods specifically tailored to sequential BOED—a direction largely unexplored by the SBI community.
It also points to a need for work on improving policy-based BOED approaches, especially for critical data acquisition settings.
Future work should also investigate hybrid approaches combining policy-based initialization with per-trajectory refinement, improved posterior validation during sequential acquisition, and techniques to reduce the simulation burden of running the MPR-GA algorithm.

\section*{Acknowledgements}
We would like to acknowledge Zhe Zhang for providing helpful feedback on this manuscript.
S.K. and S.G. were supported by the Department of Energy, Laboratory Directed Research and Development program at SLAC National Accelerator Laboratory, under contract DE-AC02-76SF00515. M.K. is supported by the US Department of Energy (DOE) under Grant No. DE-AC02-76SF00515. W.N. is supported by the National Science Foundation (NSF) under Award No. 2427856.

\section*{Impact Statement}
This paper presents work whose goal is to advance the field of Machine
Learning. There are many potential societal consequences of our work, none
which we feel must be specifically highlighted here.

\bibliography{main}
\bibliographystyle{icml2026}

%%%%%%%%%%%%%%%%%%%%%%%%%%%%%%%%%%%%%%%%%%%%%%%%%%%%%%%%%%%%%%%%%%%%%%%%%%%%%%%
%%%%%%%%%%%%%%%%%%%%%%%%%%%%%%%%%%%%%%%%%%%%%%%%%%%%%%%%%%%%%%%%%%%%%%%%%%%%%%%
% APPENDIX
%%%%%%%%%%%%%%%%%%%%%%%%%%%%%%%%%%%%%%%%%%%%%%%%%%%%%%%%%%%%%%%%%%%%%%%%%%%%%%%
%%%%%%%%%%%%%%%%%%%%%%%%%%%%%%%%%%%%%%%%%%%%%%%%%%%%%%%%%%%%%%%%%%%%%%%%%%%%%%%
\newpage
\appendix
\onecolumn

\section{Differentiable Simulators and Gradient Estimation}
\label{app:differentiable_simulators}

Our approach relies on gradient-based optimization of the EIG with respect to designs $\xi$. Computing these gradients requires backpropagating through an expected value, namely
\begin{equation}
    \nabla_\xi \mathrm{EIG}(\xi) = \nabla_\xi \mathbb{E}_{p(\theta|D)p(y|\theta,\xi)} \left[ \log \frac{p(y|\theta,\xi)}{p(y|\xi)} \right].
\end{equation}
This gradient can be estimated using either distributional gradients~\cite{reinforce} or pathwise gradients.

Distributional gradient estimators treat the expectation as an integral and differentiate through the density:
\begin{equation}
\nabla_\xi \mathbb{E}_{p(y|\theta,\xi)}[f(y)] = 
\mathbb{E}_{p(y|\theta,\xi)}\left[f(y) \nabla_\xi \log p(y|\theta,\xi)\right].
\end{equation}
These estimators work for arbitrary simulators but suffer from high variance, often requiring variance reduction techniques such as control variates~\cite{JMLR:v21:19-346}. For implicit simulators where $p(y|\theta,\xi)$ is intractable, such as those we consider in this work, computing $\nabla_\xi \log p(y|\theta,\xi)$ is not possible without additional approximations.

Pathwise gradient estimators instead differentiate through samples. When the simulator can be expressed as $y = g(\theta, \xi, \epsilon)$ where $\epsilon \sim p(\epsilon)$ is a source of randomness independent of $\xi$, we can write
\begin{equation}
\nabla_\xi \mathbb{E}_{p(y|\theta,\xi)}[f(y)] = 
\mathbb{E}_{p(\epsilon)}\left[\nabla_\xi f(g(\theta, \xi, \epsilon))\right].
\end{equation}
This gradient flows directly through the simulator's computational graph using automatic differentiation. Pathwise estimators typically have much lower variance than distributional estimators and do not require explicit knowledge of $p(y|\theta,\xi)$.

Our work assumes differentiable simulators that enable pathwise gradient estimation. This assumption of pathwise gradient compatible simulators is becoming increasingly realistic as differentiable simulators become more prevalent in scientific computing. Recent examples include particle physics simulations~\cite{neutrino-diffable,MadJax} and analysis~\cite{Neos}. 

For non-differentiable simulators, our framework can still be applied by replacing pathwise gradients with distributional gradient estimators or gradient-free optimization methods. However, this would require significant modifications to the MPR-GA algorithm and likely reduce computational efficiency. Developing efficient BOED methods for non-differentiable simulators remains an important direction for future work.

\section{Direct EIG estimator with NLE}
\label{app:direct_eig_estimate}
Consider approximating the EIG using variational distributions for both the marginal and conditional likelihoods. Let $q_0(y|\xi)$ be an approximation to the marginal likelihood $p(y|\xi)$, and let $r_\phi(y|\theta,\xi)$ be an approximation to the conditional likelihood $p(y|\theta,\xi)$.

The true EIG is defined as
\begin{equation}
\mathrm{EIG}(\xi) = \mathbb{E}_{p(\theta|D)p(y|\theta,\xi)}\left[\log p(y|\theta,\xi) - \log p(y|\xi)\right].
\label{eq:eig_true}
\end{equation}

A natural plug-in estimator using variational approximations would be
\begin{equation}
\mathcal{L}(\xi) = \mathbb{E}_{p(\theta|D)p(y|\theta,\xi)}\left[\log r_\phi(y|\theta,\xi) - \log q_0(y|\xi)\right].
\label{eq:eig_plugin}
\end{equation}

To understand the relationship between this estimator and the true EIG, we decompose the difference:
\begin{align}
\mathrm{EIG}(\xi) 
&= \mathbb{E}_{p(\theta|D)p(y|\theta,\xi)}\left[\log p(y|\theta,\xi) - \log p(y|\xi)\right] \\
&= \mathcal{L}(\xi) + \mathbb{E}_{p(\theta|D)}\left[\mathrm{KL} \big(p(y|\theta,\xi) \,\Vert\, r_\phi(y|\theta,\xi)\big)\right] - \mathrm{KL} \big(p(y|\xi) \,\Vert\, q_0(y|\xi)\big).
\label{eq:eig_bias_decomp}
\end{align}
As both KL terms are non-negative, Eq.~\eqref{eq:eig_bias_decomp} demonstrates that $\mathcal{L}(\xi)$ is neither a guaranteed upper bound nor a guaranteed lower bound on the true EIG. Instead, it differs from the true value by the difference of two non-negative terms. The first term, $\mathbb{E}_{p(\theta|D)}\left[\mathrm{KL} \big(p(y|\theta,\xi) \,\Vert\, r_\phi(y|\theta,\xi)\big)\right]$, measures how well the variational conditional $r_\phi$ approximates the true conditional likelihood, averaged over $\theta$. The second term, $\mathrm{KL} \big(p(y|\xi) \,\Vert\, q_0(y|\xi)\big)$, measures how well the variational marginal $q_0$ approximates the true marginal likelihood.

The bias of $\mathcal{L}(\xi)$ depends on the relative magnitudes of these two approximation errors. When the marginal approximation error dominates, the estimator will overestimate the true EIG. Conversely, when the conditional approximation error dominates, it will underestimate the true EIG. The estimator provides a good approximation when both errors are small or approximately equal in magnitude.

In practice, if both $r_\phi$ and $q_0$ are flexible parametric families, such as normalizing flows, trained with sufficient capacity, both KL terms can be made small, yielding $\mathcal{L}(\xi) \approx \mathrm{EIG}(\xi)$. However, unlike variational lower bounds such as the Barber--Agakov bound, this estimator provides no formal guarantee on the direction of bias.

\section{Acquisition optimization}
\label{app:acq_opt}

The diversity penalty encourages separation among the selected designs $\{\boldsymbol{\xi}_1, \ldots, \boldsymbol{\xi}_n\}$ 
using a squared hinge loss on pairwise distances:
\begin{equation}
\mathcal{L}_{\text{div}}(\boldsymbol{\xi}) = w(t) \sum_{i < j} \max(0, d_{\min} - \|\boldsymbol{\xi}_i - \boldsymbol{\xi}_j\|_2)^2,
\end{equation}
where $d_{\min}$ sets a minimum separation threshold, $\|\cdot\|_2$ denotes Euclidean distance, and $w(t)$ is annealed to zero after a fixed number of steps.
For all settings designs are normalized to a $[0,1]^d$ box 
using their bounds prior to distance computation, ensuring the penalty is scale-invariant and independent of the design space coordinate system.
For all experiments we set $d_{\min}=0.01$ 

The penalty weight is annealed following
\begin{equation}
w(t) = \begin{cases}
\lambda & \text{if } t < T_{\text{anneal}} \\
0 & \text{otherwise}
\end{cases}.
\end{equation}
This schedule prevents over-aggressive regularization in early iterations (exploration phase) while allowing the acquisition objective to dominate later (exploitation phase).
For all experiments we use $\lambda=1000$.
The squared hinge formulation ensures smooth gradients only when distances violate the constraint, 
removing all diversity penalties when designs are sufficiently separated.
Gradient updates are performed with the RMSProp algorithm with a learning rate of $0.001$ for source finding, $0.01$ for pharmacokinetic models and $0.1$ for CES models.
The increasing learning rate is to account for the increasing scale of the design spaces.
The gradients of all designs are clipped to one based on the per design gradient norm.
During the optimization all methods are burned in for 1000 gradient steps, and $T_{\text{anneal}}$ is also set to 1000 for all methods.
At the end of acquisition optimization the EIG is computed across 1000 MC samples for each design separately and this EIG estimate is used to select the final design.

To characterize the impact of the diversity penalty designs were tested in the single design acquisition setting of Section~\ref{sec:ablations}.
Table~\ref{tab:ablations} shows the impact of the diversity penalty on acquisition performance across different SBI methods.

\begin{table}[h]
\centering
\caption{Acquisition methods with and without a diversity penalty. In the acquisition optimization all methods used $256$ restart. The uncertainty on the sPCE estimates is around $0.001$ for all methods.}
\label{tab:ablations}
% \adjustbox{max width=\columnwidth}{
\begin{tabular}{lcccc}
\toprule
\textbf{Method} & \textbf{No Diversity} & \textbf{Diversity} & \textbf{Timing (m)} \\
\midrule
NRE & $0.872$ & $0.881$ & $3.07$ \\
NLE Contrastive & $0.886$ & $0.881$ & $157.3$ \\
NLE & $0.870$ & $0.885$ & $27.5$ \\
NPE & $0.868$ & $0.885$ & $12.7$ \\
\bottomrule
\end{tabular}
% }
\end{table}

\section{Models}
\label{app:hyperparameters}
\subsection{Posterior model training}
All posteriors and variational models in the acquisition step are trained with the AdamW optimizer~\cite{adamw} for 100 epochs with a learning rate of $10^{-4}$ and a weight decay of $0.01$.
For training posteriors on the source finding and pharmacokinetic models 100,000 simulated samples are used to train the posterior estimators with a batch size of 1024.
For training posteriors on the CES problem one million samples are used with a batch size of 8192.

\subsection{Neural likelihood estimation}
The normalizing flows used for neural likelihood estimation (NLE) are functionally equivalent in all of the applications in which they are used, that is in posterior estimation for estimating $p(y|\xi, \theta)$ and in acquisition optimization for estimating $p(y|\xi, \theta)$ and $p(y|\xi)$.
The flows are parameterized by two rational quadratic spline layers~\cite{durkan2019neural} with $10$ bins in each layer and the knot placements parameterized by two residual blocks with $128$ hidden nodes as implemented in the nflows library~\cite{nflows_package}. 
The base distribution of all NLE flows are normal distributions.
All input observations are scaled to the range $[-3.5, 3.5]$ and the rational quadratic functions are bounded on this same domain.
All flows have fixed linear tails.

The variational approximation used for estimating $p(y | \theta, \xi)$ and $p(y|\xi)$ often need to learn both the conditional function that describes the mean of $y$ and the conditional distribution about the mean.
To allow models to straightforwardly capture this structure we add a conditional affine transformation in the normalizing flows.
This allows the models to learn arbitrary scalings and shifts in $y$ as a function of the design $\xi$ and parameters $\theta$.

\subsection{Neural posterior estimation}
The normalizing flows used for neural posterior estimation (NPE) are functionally equivalent across posterior estimation for computing $p(\theta|y)$ and acquisition optimization for computing $p(\theta|y, \xi)$. 
The input layer to the context embeddor changes depending on the application.
All flows are parameterized by two rational quadratic spline layers~\cite{durkan2019neural} with $10$ bins in each layer and the knot placements parameterized by two residual blocks with $128$ hidden nodes as implemented in the nflows library~\cite{nflows_package}. 

For all problems except the CES benchmark, the base distribution is chosen to be a standard normal distribution. For the CES problem, the base distribution is chosen to be uniform with support on $[-3.5, 3.5]$. 
This choice reflects the natural bounded support of the CES parameters elasticity $\rho \in (0,1)$ and simplex-constrained weights $\alpha$.
All input parameters $\theta$ are normalized to the range $[-3.5, 3.5]$ and the rational quadratic functions are bounded on this same domain with fixed linear tails. 
To be consistent with NLE approaches, we include a conditional affine transformation layer that allows the flow to learn arbitrary scalings and shifts in $\theta$ as a function of the observations.
For the source finding benchmark, we represent the posterior in hyperspherical coordinates rather than Cartesian coordinates. 
This transformation reduces redundancy and improves sample efficiency when modeling distributions over locations in Euclidean space.

\subsection{Neural ratio estimation}
The neural ratio estimation (NRE) approach approach estimates mutual information $I(\theta; y | \xi)$ directly using the InfoNCE bound~\cite{oord2019representationlearningcontrastivepredictive}. 
Instead of learning explicit density models, we train a critic network $T(\theta, y, \xi)$ that assigns scores to parameter-observation-design triplets.
The critic is implemented as an multilayer perceptron with two hidden layers of 256 units each, using ReLU activations. 
Input triplets are formed by concatenating $(\theta, y, \xi)$ and passing through layer normalization before the MLP. 
The critic is trained by maximizing the InfoNCE lower bound on mutual information across all designs.
For computational efficiency with large batch sizes, we implement a chunked computation variant that processes the $N \times N$ pairwise score matrix in blocks, reducing peak memory usage while maintaining exact gradients.
The same network structure is used when training classifiers for posterior estimation.
When training posterior estimates the learning rate is fixed to $0.001$ and a batch size of $1024$ is used and the model is trained for 500 epochs.
The NRE approach is much faster to train and so it can be efficiently trained for many more epochs than the flow based models.

\section{Timing information}
\label{app:timing_information}

We compare the computational cost of per-trajectory SBI-BOED methods against policy-based approaches across all benchmarks. For per-trajectory methods, we measure the time to train posterior models, sample from posteriors during acquisition optimization, and run the acquisition optimization itself. For policy-based methods, we report the total upfront training time required before deployment.

Table~\ref{tab:source_finding_timing} provides detailed timing breakdowns for the source finding benchmark, reporting results for all SBI-BOED methods from Table~\ref{tab:source_finding_results} in the main text. Notably, the wall-clock time for acquisition optimization remains approximately constant whether using $M=1$ or $M=256$ restarts due to parallel execution, while the increased exploration from multiple restarts leads to substantially better designs.

Across all benchmarks, per-trajectory and policy-based approaches have comparable total computational costs for single-trajectory experiments. For source finding (10 steps), NPE-NRE requires ${\sim}200$ minutes total versus ${\sim}180$ minutes upfront training for DAD (averaging ${\sim}18$ minutes per design). For the pharmacokinetic model (5 steps), both NPE-NRE and DAD require ${\sim}45$ minutes (averaging ${\sim}9$ minutes per design for DAD). For CES (10 steps), NPE-NRE requires ${\sim}530$ minutes total versus ${\sim}840$ minutes for RL-BOED (averaging ${\sim}84$ minutes per design).

While policy-based methods amortize training costs across multiple trajectories, making them advantageous when many experimental runs are needed, per-trajectory methods are competitive for expensive scientific experiments where few trajectories are performed and the ability to adapt to observations justifies the per-step computational cost.

\begin{table}
\caption{The amount of time it takes each approach to train a posterior in minutes, to sample from the posterior in seconds, and to run acquistion optimization for the source finding problem.}
\centering
\adjustbox{max width=\columnwidth}{
\begin{tabular}{lccc}
\toprule
Method & Posterior Train (m) & Sampling (s) & Acq. Opt (m) \\
\midrule
\multicolumn{4}{l}{\textit{Per-trajectory M = 1 restart}} \\
NRE & 27 & 300 & 9 \\
\midrule
\multicolumn{4}{l}{\textit{Per-trajectory M = 256 restarts}} \\
NRE & 27 & 300 & 9 \\ 
NLE & 25 & 3600 & 35 \\
NPE & 10 & 1 & 16 \\
NPE-NRE & 10 & 1 & 9 \\
\bottomrule
\end{tabular}
}
\label{tab:source_finding_timing}
\end{table}

\section{Benchmark problems}
\label{app:benchmark_problems}
This section provides detailed descriptions of the benchmark problems used in our experiments, including the forward models, prior distributions, and experimental setups.

\subsection{Source Finding Benchmark}
\paragraph{Forward Model}
The source finding problem~\cite{dad} models intensity measurements $y \in \mathbb{R}$ at detector locations $\xi \in \mathbb{R}^d$ arising from $n$ unknown point sources located at positions $\theta = (\theta_1, \ldots, \theta_n)$ where $\theta_i \in \mathbb{R}^d$. The measurement model is:
\begin{equation}
y = G(\xi, \theta) + \epsilon, \quad \epsilon \sim \mathcal{N}(0, \sigma^2)
\end{equation}
where the deterministic forward map $G$ models signal decay as an inverse squared distance:
\begin{equation}
G(\xi, \theta) = \log\left(\gamma + \sum_{i=1}^{n} \frac{\alpha}{\alpha + \|\xi - \theta_i\|_2^2}\right)
\end{equation}
Here $\gamma = 0.1$ is the baseline signal strength, $\alpha = 10^{-4}$ controls the decay rate, and $\sigma = 0.5$ is the observation noise scale.

\paragraph{Prior Distribution}
To break the permutation symmetry inherent in multi-source problems, we employ an \emph{ordered prior} that constrains sources by their distance from the origin. Let $\theta^* = (\theta_1^*, \ldots, \theta_n^*)$ denote the sources sorted by radial distance
\begin{equation}
\|\theta_1^*\|_2 \leq \|\theta_2^*\|_2 \leq \cdots \leq \|\theta_n^*\|_2
\end{equation}
The ordered prior is constructed from an exchangeable base distribution
\begin{equation}
p(\theta) = n! \cdot \prod_{i=1}^{n} \mathcal{N}(\theta_i^* \mid 0, I_d) \cdot \mathbb{I}[\text{ordered}],
\end{equation}
where $\mathbb{I}[\text{ordered}]$ is an indicator function that equals 1 if the sources satisfy the radial ordering constraint and 0 otherwise. The factor $n!$ accounts for the reduced support that arises from selecting one ordering out of $n!$ permutations. 
This construction maintains an exchangeable base prior while ensuring identifiability through the ordering constraint.

\paragraph{Experimental Setup} Designs $\xi$ are optimized over the domain $[-6, 6]^2$ to maximize expected information gain.
The first measurement is always made at the origin, this was selected using a single run of the NRE acquisition optimization using the prior.

\subsection{Pharmacokinetic Benchmark}

\paragraph{Forward Model}
The pharmacokinetic problem~\cite{idad,ryan2014towards} models drug concentration measurements $y \in \mathbb{R}_+$ at time points $\xi \in [0, 24]$ (hours) following oral administration. The model follows a one-compartment absorption-elimination process with parameters $\theta = (k_a, k_e, V)$ representing the absorption rate, elimination rate, and volume of distribution respectively. The measurement model is:
\begin{equation}
y = G(\xi, \theta) + \epsilon, \quad \epsilon \sim \mathcal{N}(0, \sigma^2(\xi, \theta))
\end{equation}
where the deterministic forward map $G$ describes the plasma concentration over time:
\begin{equation}
G(\xi, \theta) = \frac{D}{V} \cdot \frac{k_a}{k_a - k_e} \cdot \left(e^{-k_e \xi} - e^{-k_a \xi}\right)
\end{equation}
Here $D = 400$ mg is the administered dose. The observation noise is heteroskedastic:
\begin{equation}
\sigma^2(\xi, \theta) = (\epsilon_{\text{rel}} \cdot G(\xi, \theta))^2 + \nu^2
\end{equation}
with relative error $\epsilon_{\text{rel}} = 0.1$ and absolute error $\nu = \sqrt{0.1}$. This captures both proportional measurement error (scaling with concentration) and constant background noise.

\paragraph{Prior Distribution}
Parameters are modeled in log-space to ensure positivity. Let $\tilde{\theta} = (\log k_a, \log k_e, \log V)$. The prior is:
\begin{equation}
p(\tilde{\theta}) = \mathcal{N}\left(\tilde{\theta} \mid \mu, \Sigma\right)
\end{equation}
where the mean and covariance are:
\begin{equation}
\mu = \begin{pmatrix} \log(1.0) \\ \log(0.1) \\ \log(20.0) \end{pmatrix}, \quad
\Sigma = 0.05 \cdot I_3
\end{equation}
These hyperparameters reflect typical physiological values: absorption occurs faster than elimination ($k_a > k_e$), with a moderate volume of distribution.

\paragraph{Experimental Setup}
Following the iDAD benchmark \citep{idad}, we conduct experiments with measurement times $\xi \in [0, 24]$ hours. 
Initial measurements are taken at $\xi_1 = 17.56$ hours always, this was selected using a single run of the NRE acquisition optimization using the prior.

\paragraph{Static baseline}
The static baseline we find makes measurements at 17.56, 0.3223 and 5.397 hours.
The static baseline over five measurements repeats the first two measurements.
These three design parameters were found by averaging the time ordered first three measurements from 100 runs of NPE-NRE data acquisition procedure.

\subsection{Constant Elasticity of Substitution (CES) Benchmark}

\paragraph{Forward Model}
The constance elasticity of substitution (CES)~\cite{foster_unified} problem models pairwise preference ratings $y \in [\epsilon, 1-\epsilon]$ over commodity bundles. Given two bundles $\xi = (\xi^{(1)}, \xi^{(2)}) \in \mathbb{R}_+^6$ (each bundle contains 3 commodities), the model predicts a preference rating based on utility differences. The parameters are $\theta = (\rho, \alpha_1, \alpha_2, \log u)$ where $\rho \in (0,1)$ controls elasticity of substitution, $(\alpha_1, \alpha_2)$ are commodity weights with $\alpha_3 = 1 - \alpha_1 - \alpha_2$ enforcing the simplex constraint, and $u = \exp(\log u)$ scales the utility. The measurement model is
\begin{equation}
y = \sigma(\eta), \quad \eta \sim \mathcal{N}(\mu_\eta, \sigma_\eta^2),
\end{equation}
where $\sigma(\cdot)$ is the sigmoid function mapping to $[0,1]$, and the mean is determined by the utility difference
\begin{equation}
\mu_\eta = u \cdot (U(\xi^{(1)}) - U(\xi^{(2)})).
\end{equation}
The CES utility function for bundle $x = (x_1, x_2, x_3)$ is given by
\begin{equation}
U(x) = \left(\sum_{i=1}^{3} \alpha_i x_i^\rho\right)^{1/\rho}.
\end{equation}
The observation noise is heteroskedastic and depends on bundle dissimilarity:
\begin{equation}
\sigma_\eta = u \cdot \sigma_0 \cdot (1 + \|\xi^{(1)} - \xi^{(2)}\|_2)
\end{equation}
where $\sigma_0 = 0.005$ is the base noise scale. Responses are censored to $[\epsilon, 1-\epsilon]$ with $\epsilon = 2^{-22}$ to avoid numerical issues at the sigmoid boundaries.

\paragraph{Prior Distribution}
The prior factorizes over parameter types:
\begin{align}
\rho &\sim \text{Beta}(1, 1) = \text{Uniform}(0, 1) \\
(\alpha_1, \alpha_2, \alpha_3) &\sim \text{Dirichlet}(1, 1, 1) \\
\log u &\sim \mathcal{N}(1, 9)
\end{align}
The Dirichlet prior ensures weights sum to unity, while the log-normal prior on $u$ ensures positive utility scaling. The uniform prior on $\rho$ encompasses the full range from perfect substitutes ($\rho \to 1$) to Leontief preferences ($\rho \to 0$).

\paragraph{Experimental Setup}
Designs $\xi \in [0, 100]^6$ consist of two bundles with three commodities each. The initial measurement is taken at the bundle pair
\begin{equation}
\xi_0 = (24.9, 52.0, 81.8, 53.5, 62.5, 35.5),
\end{equation}
which was selected using a single run of the NRE acquisition optimization using the prior.

\section{Implementation and Software}

Our approach was implemented in Python using the PyTorch deep learning framework \cite{torch_two} 
and the Simulation-Based Inference (SBI) library \cite{BoeltsDeistler_sbi_2025}. Neural network architectures were implemented using nflows for normalizing flow models~\cite{nflows_package}. Numerical computations were performed with NumPy \cite{numpy} and SciPy \cite{scipy}, while visualization was handled with Matplotlib \cite{Hunter_Matplotlib_A_2D_2007}. 
Experiment configuration was managed using Hydra \cite{hydra}, and workflow orchestration was performed with Snakemake \cite{snakemake}. The computational environment was managed using Pixi \cite{pixi_Reproducible, Arts_pixi}.

%%%%%%%%%%%%%%%%%%%%%%%%%%%%%%%%%%%%%%%%%%%%%%%%%%%%%%%%%%%%%%%%%%%%%%%%%%%%%%%
%%%%%%%%%%%%%%%%%%%%%%%%%%%%%%%%%%%%%%%%%%%%%%%%%%%%%%%%%%%%%%%%%%%%%%%%%%%%%%%

\end{document}